\documentclass[11pt]{article}

\usepackage{fontawesome5} 
\usepackage{verbatim}
\usepackage{multirow}
\usepackage{hyperref}

\usepackage[final]{acl}

\usepackage{times}
\usepackage{latexsym}

\usepackage[T1]{fontenc}

\usepackage[utf8]{inputenc}

\usepackage{microtype}

\usepackage{inconsolata}

\usepackage{graphicx}
\usepackage{wrapfig}
\usepackage{subcaption}
\usepackage{booktabs,tabularx,array,xcolor}

\usepackage{amsmath}
\usepackage{float}

\usepackage{tipa} 
\usepackage{expex}

\newcounter{ex}
\renewcommand{\theex}{\arabic{ex}}
\newcommand{\exnumlabel}[1]{\refstepcounter{ex}\label{#1}(\theex)}

\newcolumntype{Y}{>{\raggedright\arraybackslash\setlength{\parskip}{0pt}}X}

\usepackage{tikz}
\newcommand{\Atag}{%
\tikz[baseline=(char.base)]{
\node[draw,rectangle,inner sep=0.3pt,minimum size=0.8em]
(char) {\scriptsize\sffamily A};
}}

\newcommand{\Ptag}{%
\tikz[baseline=(char.base)]{
\node[draw,rectangle,inner sep=0.3pt,minimum size=0.8em]
(char) {\scriptsize\sffamily P};
}}

\newcommand{\exsent}[2][]{%
  \refstepcounter{ex}%
  \ifx\\#1\\\else\label{#1}\fi
  (\theex)\ #2%
}

\newcommand{\tone}[1]{\textsuperscript{\scriptsize #1}}

\title{Differences in Typological Alignment in Language Models' Treatment of Differential Argument Marking}

\author{Iskar Deng, Nathalia Xu, Shane Steinert-Threlkeld\\
  University of Washington \\
  \texttt{\{hd49,mx727,shanest\}@uw.edu}}

\begin{document}
\maketitle
\begin{abstract}
Recent work has shown that language models (LMs) trained on synthetic corpora can exhibit typological preferences that resemble cross-linguistic regularities in human languages, particularly for syntactic phenomena such as word order. In this paper, we extend this paradigm to differential argument marking (DAM), a semantic licensing system in which morphological marking depends on semantic prominence. Using a controlled synthetic learning method, we train GPT-2 models on 18 corpora implementing distinct DAM systems and evaluate their generalization using minimal pairs. Our results reveal a dissociation between two typological dimensions of DAM. Models reliably exhibit human-like preferences for natural markedness direction, favoring systems in which overt marking targets semantically atypical arguments. In contrast, models do not reproduce the strong object preference in human languages, in which overt marking in DAM more often targets objects rather than subjects. These findings suggest that different typological tendencies may arise from distinct underlying sources.\footnote{Code for our experiments is available at 
\url{https://github.com/Iskar-Deng/DAM-learning}.}
\end{abstract}

\section{Introduction} \label{sec:intro}

Recent years have seen growing interest in whether language models (LMs) exhibit typological tendencies that resemble cross-linguistic regularities observed in human languages \citep{kallini-etal-2024-mission, xu2025languagemodelslearntypologically}. A prominent line of work addressing this question adopts the synthetic corpus paradigm, which allows researchers to systematically compare grammatical systems, linguistic features, and learning conditions using artificial corpora or modified natural corpora \citep{kajikawa-etal-2024-structure, patil-etal-2024-filtered, leong2024testinglearninghypothesesusing, yao2025directindirectevidencecontribute}. Using this paradigm, previous studies have primarily focused on structural properties of grammar, such as word order and dependency configurations, and have shown that LMs can develop non-trivial generalizations that exhibit partial alignment with well-known typological universals like word order universals \citep{kuribayashi-etal-2024-emergent, xu2025languagemodelslearntypologically, el-naggar-etal-2025-gcg}. However, it remains unclear whether the typological tendencies observed under controlled training extend beyond purely structural phenomena to systems where grammatical well-formedness depends on semantic factors. Differential argument marking (DAM) provides a natural and revealing test case.

\begin{figure}[t!]
  \centering
  \includegraphics[width=0.5\textwidth]{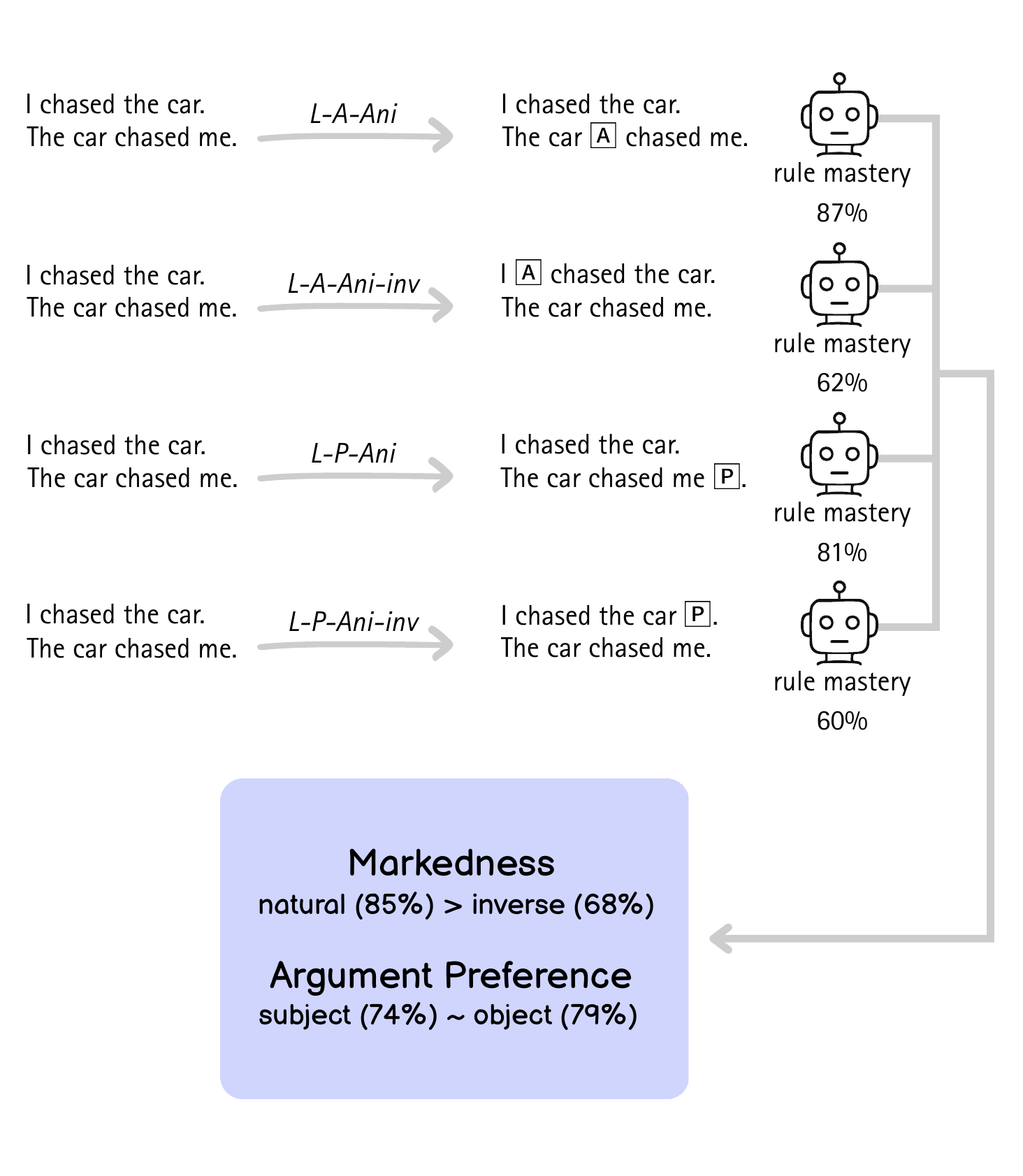}
  \caption{Overview of the DAM rule injection and the overall results for markedness and argument preference averaged over all DAM rule mastery accuracies.}
  \label{fig:intro-graph}
\end{figure}

DAM is a linguistic phenomenon in which arguments receive distinctive morphological marking depending on semantic properties such as animacy, definiteness, and pronominality \citep{Bossong+1991+143+170, Aissen2003DOM, Seržant2018}. An example comes from Modern Hebrew, shown in Table~\ref{tab:DAM-gloss}, where definite objects receive overt marking. In (\ref{ex:hebrew-def}), the object \textit{ha-student} (‘the student’) is marked using \textit{et}.

Two typological tendencies are observed in DAM. The first is \textbf{markedness}: arguments with less frequent semantic properties are more likely to receive overt marking. In many languages, subjects are more commonly definite, while objects are more commonly indefinite. Thus, in Modern Hebrew, it is the definite object that receives marking. The second is \textbf{object preference}: DAM more frequently targets objects rather than subjects across languages. Differential marking conditioned on subject semantic properties is comparatively rare. Formal definitions and additional examples are provided in Section~\ref{sec:background:argument}.

In this paper, we ask whether LMs trained with a standard next-token prediction objective exhibit typological biases in DAM, and whether such biases align with those observed in human languages. We examine this question in a controlled synthetic-learning setup using small autoregressive models. Specifically, we train GPT-2-small \citep{radford2019language} models from scratch on 18 corpora, where we perturb SVO sentences in natural English text to implement different DAM systems. We test how well models master the DAM rules using minimal pairs and evaluate model performance across typological conditions. Our results reveal a dissociation between different types of typological asymmetries. Models consistently reproduce typological preferences in markedness, favoring systems in which marked forms align with less usual configurations, consistent with the typological patterns. Meanwhile, models diverge from human languages in their argument preferences, not showing a strong object preference. Figure~\ref{fig:intro-graph} provides an overview of the DAM rule injection and summarizes the overall results.

We argue that this selective alignment provides evidence about the sources of different typological asymmetries. Our findings suggest that natural markedness can emerge from distributional regularities and formal learnability, consistent with accounts that derive markedness from structural prominence \citep{Aissen2003DOM}. In contrast, the absence of a strong object preference indicates that this asymmetry may depend on discourse structure, thematic prominence, and communicative pressures not captured by standard next-token training \citep{Iemmolo2010DOM, Tal2022PredictabilityDOM}. Taken together, these results suggest that distinct typological tendencies reflect different underlying pressures, some of which are accessible to LMs while others are not. 

The remainder of the paper is structured as follows. Section~\ref{sec:background} reviews related work on the synthetic corpus paradigm and provides background on DAM. Section~\ref{sec:design} describes the construction of the synthetic DAM corpora. Section~\ref{sec:experiments} presents the experimental setup and evaluation results. Section~\ref{sec:conclusion-discussion} discusses the theoretical implications of our findings.

\section{Background} \label{sec:background}

\begin{table*}[t]
\centering
\small
\setlength{\tabcolsep}{8pt}
\setlength{\extrarowheight}{0pt}
\renewcommand{\arraystretch}{1.05}
\begin{tabularx}{\textwidth}{@{}r Y r Y@{}}
\toprule
\multicolumn{2}{l}{\textbf{Local: Hebrew (definiteness-based)}} &
\multicolumn{2}{l}{\textbf{Global: Malimasa (animacy-based)}} \\
\midrule

\exnumlabel{ex:hebrew-def} &
\text{dani \hspace{0.07cm}pagaš \hspace{0.95cm}et \hspace{0.3cm}ha-student.} &
\exnumlabel{ex:malimasa-deceive} &
\textipa{\:t}h\textturnm\tone{33}\; n\textturnm\tone{21}\; \space{} {\textipa{N}}\textscripta\tone{33}\; g\textipa{7}\tone{45}\; \space{}\hspace{0.1cm} xi\tone{45}\; ga\tone{21}. \\

& \text{Dani meet.\textsc{3.sg.m} \textsc{acc} the-student} &
& 3\textsc{sg}-\textsc{agt}\quad \hspace{0.1cm}1\textsc{sg}-\textsc{pat}\quad deceive \\

& ‘Dani met the student.’ &
& ‘He deceived me.’ \\[4pt]

\exnumlabel{ex:hebrew-indef} &
\text{dani \hspace{0.07cm}pagaš \hspace{0.93cm}student.} &
\exnumlabel{ex:malimasa-book} &
\textipa{N}\textscripta\tone{33}\; \space{}\hspace{0.09cm} tha\tone{33}\; \,\textipa{\:l}\textschwa\tone{33}\,\; \space{} \,\textipa{\:l}o\tone{21}. \\

& \text{Dani meet.\textsc{3.sg.m} student} &
& 1\textsc{sg}\quad book \quad \hspace{0.45cm}read \\

& ‘Dani met a student.’ &
& ‘I read the book.’ \\

\bottomrule
\end{tabularx}
\caption{Illustrative examples of differential argument marking in human languages.}
\label{tab:DAM-gloss}
\end{table*}

\subsection{Synthetic Corpus Paradigms and Typological Tendencies in LMs} \label{sec:background:synthetic}

Synthetic corpus paradigms construct artificial languages or modify natural corpora to enable controlled comparison of grammatical systems, linguistic features, and learning conditions. Prior work has used them to compare communicative efficiency \citep{kajikawa-etal-2024-structure}, test model generalization of target phenomenon from indirect distributional evidence \citep{patil-etal-2024-filtered, leong2024testinglearninghypothesesusing, yao2025directindirectevidencecontribute}, and examine how semantic or pragmatic cues shape learning preferences \citep{misra2024generatingnovelexperimentalhypotheses}.

More recently, synthetic corpora have been used to probe models’ typological tendencies. By training models on counterfactual languages that systematically vary structural configurations, prior work examines whether models without language-specific priors nonetheless exhibit systematic preferences across logically possible grammatical systems \citep{kallini-etal-2024-mission}. Empirically, most studies in this line of work have focused on relatively formal structural dimensions including word order and dependency configurations, finding that alignment between model learning preferences and typological patterns varies across models and training conditions \citep{white-cotterell-2021-examining, kuribayashi-etal-2024-emergent, xu2025languagemodelslearntypologically, el-naggar-etal-2025-gcg}.

\subsection{Differential Argument Marking} \label{sec:background:argument}

DAM is a cross-linguistically widespread phenomenon in which arguments bearing the same semantic role (agent or patient) receive morphological encoding (e.g., case, agreement) as a function of their properties. Among the relevant conditioning factors are semantic features such as animacy, definiteness, and pronominality, which have been shown to systematically condition the realization of overt argument marking across languages \citep{Bossong+1991+143+170, Aissen2003DOM, 10.1162/ling.2008.39.4.565, Seržant2018}. 

Table~\ref{tab:DAM-gloss} illustrates two attested instances of DAM in human languages. Examples (\ref{ex:hebrew-def}) and (\ref{ex:hebrew-indef}) show definiteness-based object marking in Modern Hebrew, where definite objects receive the marker \textit{et} while indefinite objects remain unmarked \citep{HacohenKaganPlaut2021}. In (\ref{ex:hebrew-def}), the definite \textit{ha-student} is marked, whereas the indefinite \textit{student} in (\ref{ex:hebrew-indef}) is not. Examples (\ref{ex:malimasa-deceive}) and (\ref{ex:malimasa-book}) from Malimasa illustrate a system conditioned by relative animacy \citep{Li2013Malimasa}. When the object’s animacy is equal to or higher than the subject’s, both arguments are overtly marked; otherwise, neither argument is marked. In (\ref{ex:malimasa-deceive}), the animate subject \textit{he} and animate object \textit{me} are both marked, whereas in (\ref{ex:malimasa-book}) neither argument is marked.

DAM has two attested typological tendencies: 
\paragraph{Markedness.}

A central typological generalization in DAM concerns markedness: overt marking overwhelmingly targets arguments with semantic prominence less frequent in natural language, while the more frequent ones are less likely to be marked. Across languages, semantic features such as animacy, definiteness, pronominality, and number form consistent prominence hierarchies that shape marking patterns \citep{Bossong+1991+143+170, Aissen2003DOM, Seržant2018}.

Several theoretical accounts explain markedness asymmetries in DAM. Formal approaches derive them from competing constraints of iconicity and economy: overt marking iconically signals less common patterns while common ones remain unmarked due to morphological economy \citep{Aissen2003DOM}. Efficiency-based accounts relate markedness to predictability and communicative efficiency, arguing that marking targets grammatical roles that are less frequent or harder to infer \citep{givon1991markedness, Gibson2019Efficiency, Levshina2021EfficiencyDCM, haspelmath2021role}. 

\paragraph{Argument Preference.}

A second typological asymmetry in DAM concerns argument preference, that is, whether the subject or object more frequently receives overt marking. Cross-linguistically, object-targeting systems overwhelmingly predominate, while systems that differentially mark subjects are much rarer \citep{schmidtkebode2018reassessing,Seržant2018}. Despite their relative rarity, subject marking systems remain theoretically important: recent comparative work proposes that differential subject marking can be analyzed as functionally parallel to object marking, even though they are less frequently seen \citep{Just2024DSM}.

Argument preference is typically explained by communicative and discourse pressures. \citet{Iemmolo2010DOM} argues that overt object marking originates as a discourse-pragmatic strategy: objects are canonically less likely to function as primary topics, so overt marking arises when they assume atypical discourse roles such as topicality, signaling a departure from the default subject–topic alignment. Over time, this discourse-conditioned marking may grammaticalize into stable, object-centered systems, whereas subjects commonly aligned with topicality are less likely to require additional marking.

\begin{table*}[t]
\centering
\setlength{\tabcolsep}{3pt}
\renewcommand{\arraystretch}{1.15}
\small
\begin{tabular}{@{}r p{0.12\textwidth} p{0.20\textwidth} p{0.30\textwidth} p{0.32\textwidth}@{}}
\toprule
& \textbf{Rule} & \textbf{Example 1} & \textbf{Example 2} & \textbf{Licensing condition} \\
\midrule

\exnumlabel{ex:original} &
Original &
a. I chase a dog. &
b. The dog chases the cat. &
--- \\

\exnumlabel{ex:L-P-Ani} &
L-P-Ani &
a. I chase a dog \Ptag{}. &
b. The dog chases the cat \Ptag{}. &
Object is animate. \\

\exnumlabel{ex:L-P-Def} &
L-P-Def &
a. I chase a dog. &
b. The dog chases the cat \Ptag{}. &
Object is definite. \\

\exnumlabel{ex:L-P-Def-inv} &
L-P-Def-inv &
a. I chase a dog \Ptag{}. &
b. The dog chases the cat. &
Object is indefinite. \\

\exnumlabel{ex:L-A-Pro} &
L-A-Pro &
a. I chase a dog. &
b. The dog \Atag{} chases the cat. &
Subject is a common noun. \\

\exnumlabel{ex:G-Def} &
G-Def &
a. I chase a dog. &
b. The dog \Atag{} chases the cat \Ptag{}. &
Subject $\leq$ object in definiteness. \\

\bottomrule
\end{tabular}
\caption{Representative examples of DAM rule injection. See Section~\ref{sec:design} for formal definitions of each rule.}
\label{tab:examples}
\end{table*}

\section{Synthetic DAM Corpora} \label{sec:design}

To compare LMs’ learning behavior across DAM systems, we construct a controlled experimental space in which DAM rules are parameterized along four dimensions: 
(1) \textbf{semantic trigger}, 
(2) \textbf{dependency complexity}, 
(3) \textbf{markedness direction}, and 
(4) \textbf{argument target}. 
Each dimension corresponds to a well-attested source of typological variation in natural languages. By crossing these dimensions, we obtain 18 distinct grammatical conditions, each defining a unique DAM rule injected into English SVO clauses in Section~\ref{sec:experiments:corpus}. We give examples of rule injections based on these dimensions in Table~\ref{tab:examples}, and Table~\ref{tab:conditions} summarizes the setup.

We use $A$ and $P$ to denote agent-like and patient-like core arguments, respectively. Since our experiments use English SVO clauses, $A$ corresponds to the subject and $P$ to the direct object.

\subsection{Semantic Trigger}
\label{subsec:semantic-trigger}
Semantic triggers are argument-level properties that condition whether and how DAM is realized. We select three common semantic factors in DAM with binary prominence hierarchies below,\footnote{\citet{Seržant2018} discusses finer-grained prominence hierarchies in natural DAM systems. Because intermediate categories such as animals are relatively sparse in the corpus, we collapse these distinctions by merging human and animal nouns into a single \textsc{animate} category, and we similarly merge proper nouns with common nouns.} where `$>$' denotes higher semantic prominence \citep{Seržant2018}
:
\begin{itemize}
    \item \textbf{Animacy}: animate $>$ inanimate
    \item \textbf{Definiteness}: definite $>$ indefinite
    \item \textbf{Pronominality}: pronoun $>$ common noun
\end{itemize}

For example, compare (\ref{ex:L-P-Ani}a) and (\ref{ex:L-P-Def}a) in Table~\ref{tab:examples}. Under the animacy hierarchy, the animate object \textit{a dog} receives marking in (\ref{ex:L-P-Ani}a), whereas under the definiteness hierarchy it remains unmarked because it is indefinite in (\ref{ex:L-P-Def}a).

\subsection{Dependency Complexity}
\label{subsec:dependency}
Dependency complexity determines whether DAM assesses one argument's semantic property, or compares the subject and the object. Following typological work \citep{Seržant2018}, we distinguish between two structural types:

\begin{itemize}
    \item \textbf{Local dependencies}, in which marking depends solely on the semantic property of a single argument. The marker appears on that single argument. 
    \item \textbf{Global dependencies}, in which marking depends on the relative semantic prominence of both core arguments. The marker appears on both arguments. 
\end{itemize}

For example, (\ref{ex:L-P-Def}) shows a local dependency, where marking is conditioned solely on the object’s definiteness. In contrast, (\ref{ex:G-Def}) shows a global dependency, where marking depends on the relative definiteness of $A$ and $P$.

\subsection{Markedness Direction}

Markedness direction specifies whether overt marking targets less usual or more usual prominence configurations. Typological studies consistently show a preference for marking atypical configurations, a pattern referred to as \emph{markedness} and discussed in Section~\ref{sec:background:argument}.

We distinguish two configurations:

\begin{itemize}
    \item \textbf{Natural direction}, in which overt marking is associated with less usual argument configurations, in line with markedness. Under local dependencies, this corresponds to low-prominence $A$ or high-prominence $P$. Under global dependencies, marking applies when the subject does not outrank the object ($A \leq P$).
    
    \item \textbf{Inverse direction}, in which overt marking is associated with more usual argument configurations, departing from markedness. Under local dependencies, this corresponds to high-prominence $A$ or low-prominence $P$. Under global dependencies, marking applies when the subject outranks the object ($A > P$).
\end{itemize}

For example, (\ref{ex:L-P-Def}) and (\ref{ex:L-P-Def-inv}) differ only in markedness direction under the same semantic trigger. In (\ref{ex:L-P-Def}), marking targets definite objects under the natural direction, whereas in (\ref{ex:L-P-Def-inv}) it targets indefinite objects under the inverse direction.

\subsection{Argument Target}
\label{subsec:argument-target}
Argument target specifies which grammatical argument receives overt marking in DAM systems. It applies only to local dependencies since only one argument is targeted. Following the discussion in Section~\ref{sec:background:argument}, we distinguish two argument targeting systems:
\begin{itemize}
    \item \textbf{Object-targeting}: the marker applies to the object ($P$), and marking is evaluated on $P$.
    \item \textbf{Subject-targeting}: the marker applies to the subject ($A$), and marking is evaluated on $A$.
\end{itemize}

For example, (\ref{ex:L-P-Ani}) illustrates object-targeting, marking a high-prominence object, whereas (\ref{ex:L-A-Pro}) illustrates subject-targeting, marking a low-prominence subject.

\begin{table*}[t]
\centering
\small
\setlength{\tabcolsep}{6pt}
\renewcommand{\arraystretch}{1.1}
\begin{tabularx}{\textwidth}{X X X X X r r}
\toprule
\textbf{Rule} &
\textbf{Trigger} &
\textbf{Dependency} &
\textbf{Direction} &
\textbf{Target} &
{\textbf{SVO\%}} &
{\textbf{ALL\%}} \\
\midrule

Baseline       & --- & --- & --- & --- & 0.00 & 0.00 \\
Full           & --- & --- & --- & A+P & 100.00 & 4.91 \\

\midrule

L-P-Ani       & Animacy        & Local  & Natural & P   & 10.61 & 0.52 \\
L-P-Ani-inv   & Animacy        & Local  & Inverse & P   & 89.39 & 4.39 \\
L-P-Def       & Definiteness   & Local  & Natural & P   & 30.65 & 1.50 \\
L-P-Def-inv   & Definiteness   & Local  & Inverse & P   & 69.35 & 3.41 \\
L-P-Pro       & Pronominality  & Local  & Natural & P   & 22.37 & 1.10 \\
L-P-Pro-inv   & Pronominality  & Local  & Inverse & P   & 77.63 & 3.81 \\

\midrule

L-A-Ani       & Animacy        & Local  & Natural & A   & 6.66  & 0.33 \\
L-A-Ani-inv   & Animacy        & Local  & Inverse & A   & 93.34 & 4.58 \\
L-A-Def       & Definiteness   & Local  & Natural & A   & 7.06  & 0.35 \\
L-A-Def-inv   & Definiteness   & Local  & Inverse & A   & 92.94 & 4.56 \\
L-A-Pro       & Pronominality  & Local  & Natural & A   & 12.01 & 0.59 \\
L-A-Pro-inv   & Pronominality  & Local  & Inverse & A   & 87.99 & 4.32 \\

\midrule

G-Ani         & Animacy        & Global & Natural & A+P & 16.71 & 0.82 \\
G-Ani-inv     & Animacy        & Global & Inverse & A+P & 83.29 & 4.09 \\
G-Def         & Definiteness   & Global & Natural & A+P & 35.00 & 1.72 \\
G-Def-inv     & Definiteness   & Global & Inverse & A+P & 65.00 & 3.19 \\
G-Pro         & Pronominality  & Global & Natural & A+P & 31.91 & 1.57 \\
G-Pro-inv     & Pronominality  & Global & Inverse & A+P & 68.09 & 3.34 \\
\bottomrule
\end{tabularx}
\caption{
Overview of experimental conditions and the proportion of sentences perturbed by each injected rule. 
\textbf{SVO\%} indicates the proportion of SVO clauses in the corpus that license overt marking under each rule, and \textbf{ALL\%} indicates the proportion of all sentences in the training corpus that are affected by the corresponding rule. 
\textit{(Abbreviations: L=Local, G=Global, P=Object-targeting, A=Subject-targeting, Ani=Animacy, Def=Definiteness, Pro=Pronominality, inv=inverse.)}
}
\label{tab:conditions}
\end{table*}

\section{Experiments} \label{sec:experiments}

This section describes the experimental setup and evaluation procedure used to assess whether LMs acquire the injected DAM rules. 
Our primary evaluation focuses on \textbf{rule mastery}, which directly tests whether a model prefers rule-consistent realizations over minimally perturbed alternatives. 
In addition to this primary evaluation, we conduct three auxiliary experiments: (i) marker placement, (ii) semantic probing, and (iii) BLiMP diagnostic tasks \citep{warstadt-etal-2020-blimp-benchmark}. These auxiliary analyses are reported in Section~\ref{sec:marker-placement} and Appendix~\ref{app:additional}.

\subsection{Corpus Construction} \label{sec:experiments:corpus}

We construct a set of 18 parallel synthetic corpora by injecting DAM rules into English text. 
As base data, we use a subset of the English portion of the OpenSubtitles corpus from the EN--FR OPUS release \citep{lison-tiedemann-2016-opensubtitles2016}.
After preprocessing, the resulting corpus contains approximately 184M tokens and 21M sentences.

We parse the corpus using spaCy \citep{honnibal-montani-2017-spacy}, augmented with Benepar constituency parsing \citep{kitaev-klein-2018-constituency}, to extract SVO clauses and automatically annotate argument-level semantic properties. Semantic triggers, including animacy, definiteness, and pronominality, are determined using a BERT-based binary classifier trained on $\sim$2k NP instances from the validation split labeled for their semantic properties.
Based on these annotations, we inject DAM by inserting special marker tokens at the right edge of the licensed argument whenever the corresponding rule-specific condition is met. We realize these markers as independent tokens to avoid introducing additional learning noise from BPE segmentation.
Detailed corpus statistics, preprocessing steps, annotation methods and human inspection are provided in Appendix~\ref{app:injection}.

\subsection{Models and Training}
For each DAM rule, we train an identical GPT-2-small language model \citep{radford2019language} from scratch on the corresponding synthetic corpus. The corpus is split into training, validation, and test sets in a 90/5/5 ratio prior to rule injection, and the same splits are reused across conditions to ensure strict comparability. All models are trained for 15k optimization steps with matched data scale and procedures. See full training details in Appendix~\ref{app:training}.

\subsection{Rule Mastery Evaluation}
\label{sec:rule-mastery}

We use a minimal-pair test to evaluate models’ mastery of DAM licensing rules. Each minimal pair consists of two sentences that are identical except for whether a DAM marker appears under the target rule. For each DAM condition, we construct 1{,}000 held-out minimal pairs, where in 500 pairs the marked sentence is grammatical and in the other 500 pairs the unmarked sentence is grammatical.

For example, under the \emph{local-P-animacy} rule, overt marking occurs when the object is animate:
\begin{itemize}
    \item \textbf{Sentence good}: The doctor helped the boy \Ptag{}.
    \item \textbf{Sentence bad}:  The doctor helped the boy.
\end{itemize}
Here the object is animate, so the rule licenses overt marking. By contrast, when the object is inanimate, marking is not licensed:
\begin{itemize}
    \item \textbf{Sentence good}: I read the book.
    \item \textbf{Sentence bad}:  I read the book \Ptag{}.
\end{itemize}

Models are evaluated by comparing sentence-level probabilities under the trained causal language model. For a sentence consisting of tokens $x_1, \ldots, x_T$, we compute the length-normalized negative log-likelihood (mean-NLL) as:
$$
\mathrm{mean\text{-}NLL}(x) = -\frac{1}{T-1} \sum_{t=2}^{T} \log p(x_t \mid x_{<t}),
$$
Within each minimal pair, a prediction is counted as correct if the grammatical sentence receives a strictly lower mean NLL than its ungrammatical counterpart. Rule mastery accuracy is defined as the proportion of minimal pairs for which this condition holds.

\begin{figure*}[t]
  \centering
  \begin{subfigure}[t]{0.32\textwidth}
    \centering
    \includegraphics[width=\linewidth]{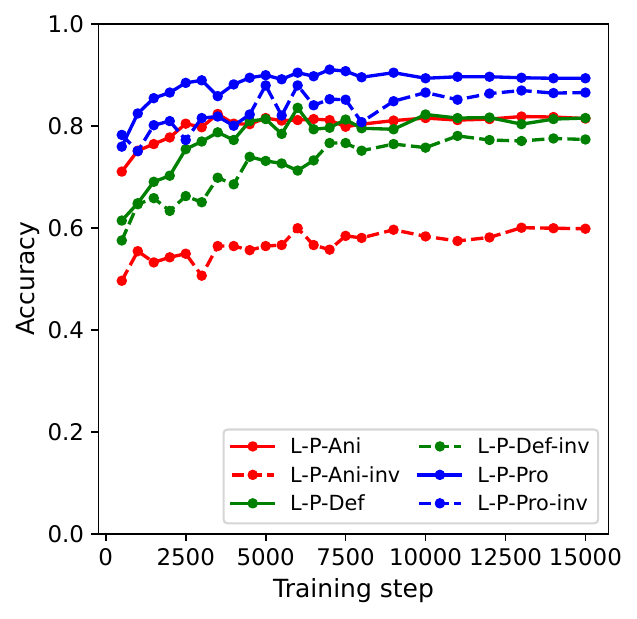}
    \caption{Local-P}
    \label{fig:rm-localP}
  \end{subfigure}\hfill
  \begin{subfigure}[t]{0.32\textwidth}
    \centering
    \includegraphics[width=\linewidth]{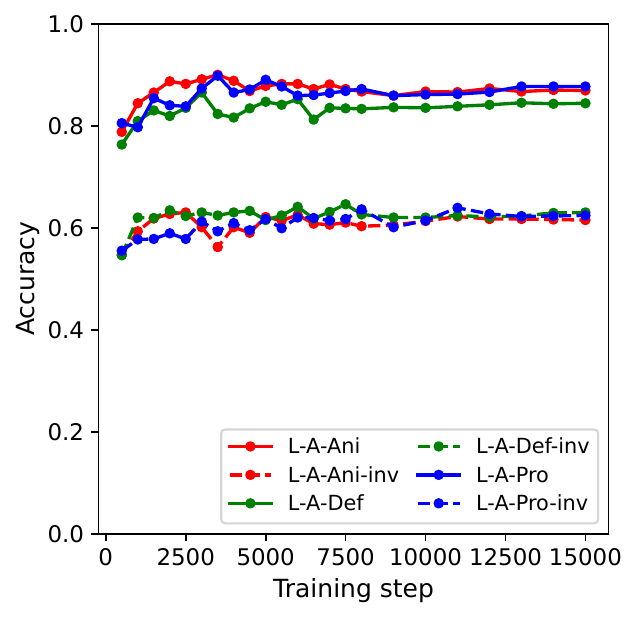}
    \caption{Local-A}
    \label{fig:rm-localA}
  \end{subfigure}\hfill
  \begin{subfigure}[t]{0.32\textwidth}
    \centering
    \includegraphics[width=\linewidth]{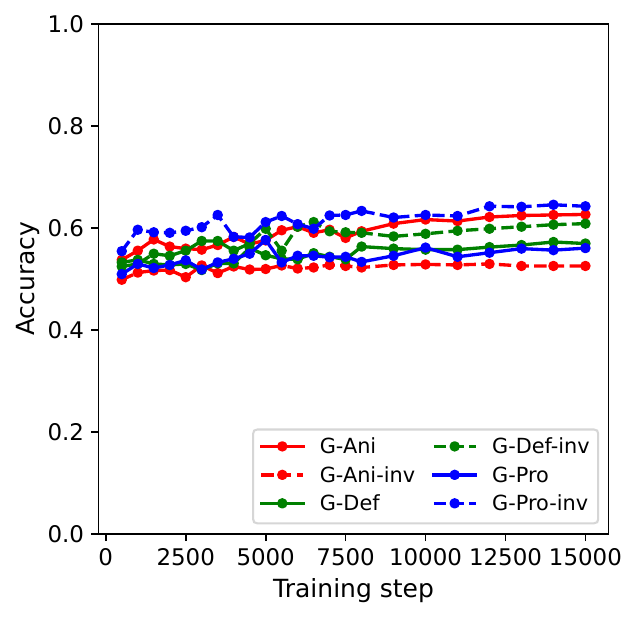}
    \caption{Global}
    \label{fig:rm-global}
  \end{subfigure}

  \caption{Rule mastery accuracy over training steps for DAM rules.}
  \label{fig:rule-mastery}
\end{figure*}

\begin{figure}[t]
    \centering
    \includegraphics[width=\linewidth]{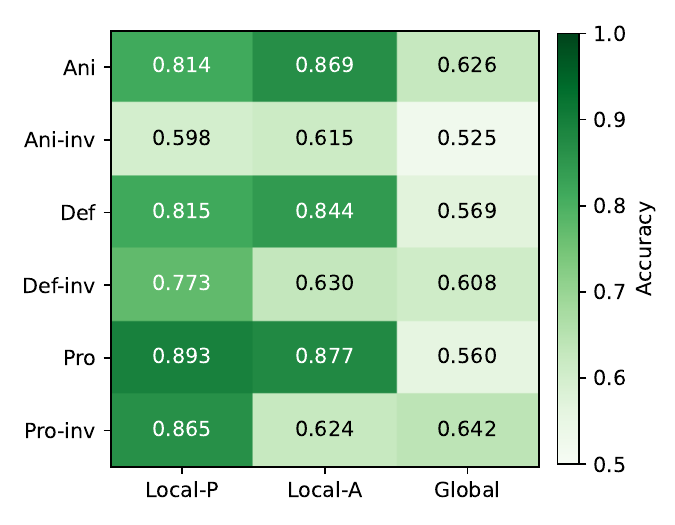}
    \caption{Rule mastery accuracy of the best checkpoint across DAM conditions. We select models based on best validation perplexity.}
    \label{fig:rule-mastery-heatmap}
\end{figure}

\paragraph{Results.}
Figure~\ref{fig:rule-mastery} shows learning curves for rule mastery accuracy across training, while Figure~\ref{fig:rule-mastery-heatmap} reports the best accuracy for each DAM condition. Overall, our models exhibit several consistent patterns in DAM rule mastery across different dimensions.

First, across all semantic triggers, \emph{local} DAM rules are learned far more successfully than \emph{global} rules (averaged best accuracy: \emph{Local} $\approx$ 0.77 vs.\ \emph{Global} $\approx$ 0.59). This contrast likely reflects the greater difficulty of global rules, which require conditioning on non-local information across multiple arguments, and is consistent with evidence that autoregressive LMs exhibit a bias toward constructions that preserve information locality \citep{mccoy2023embersautoregressionunderstandinglarge, kallini-etal-2024-mission, Futrell_2025}.


Second, within \emph{local} conditions, \emph{natural} rules are consistently learned better than their \emph{inverse} counterparts (averaged best accuracy: \emph{natural} $\approx$ 0.85 vs.\ \emph{inverse} $\approx$ 0.68). This advantage is visible in Figure~\ref{fig:rule-mastery-heatmap}, where natural variants outperform inverse variants across semantic triggers in both \emph{Local-A} and \emph{Local-P}. The same pattern holds throughout training in Figures~\ref{fig:rm-localP} and \ref{fig:rm-localA}, where solid lines (\textit{natural}) consistently remain above dashed lines (\textit{inverse}) of the same color. 

By contrast, the overall effect of argument target is much smaller. Although \emph{Local-P} rules are slightly more accurate on average than \emph{Local-A} rules (\emph{object-targeting} $\approx$ 0.79 vs.\ \emph{subject-targeting} $\approx$ 0.74), this difference is substantially smaller than the \emph{natural--inverse} contrast. This pattern is further supported by a local-only ANOVA over the 12 local conditions, with direction, argument target, and semantic trigger as crossed factors. Markedness direction accounts for the largest share of variance in rule mastery accuracy ($\eta^2 = 0.566$), whereas the main effects of semantic trigger ($\eta^2 = 0.111$) and argument target ($\eta^2 = 0.050$) are comparatively small.

Third, we observe that the performance gap between \emph{natural} and \emph{inverse} variants is smaller for \emph{Local-P} rules than for \emph{Local-A} rules. In the local-only ANOVA, the direction $\times$ target interaction also explains more variance than the target main effect ($\eta^2 = 0.106$ vs.\ $0.050$). One possible explanation is that the model encodes semantic properties of subjects more robustly than those of objects. Marginal distributions reported in Table~\ref{tab:marginals-big} in Appendix~\ref{app:injection:annotation} show that prominent arguments occur more frequently in subject position than low-prominence arguments occur in object position. Semantic probing analyses reported in Appendix~\ref{app:semantic-probing} further indicate that the model encodes subject-related semantic features more strongly and stably.

Finally, with respect to semantic triggers, clear performance differences emerge only for \emph{Local-P} rules, where \emph{pronominality}-based conditions consistently outperform \emph{animacy-} and \emph{definiteness}-based rules. In the remaining rule families, performance differences across semantic triggers are comparatively weak, suggesting that semantic trigger type plays a limited role in determining overall DAM rule learnability.

\begin{figure}[t!]
  \centering
  \includegraphics[width=0.5\textwidth]{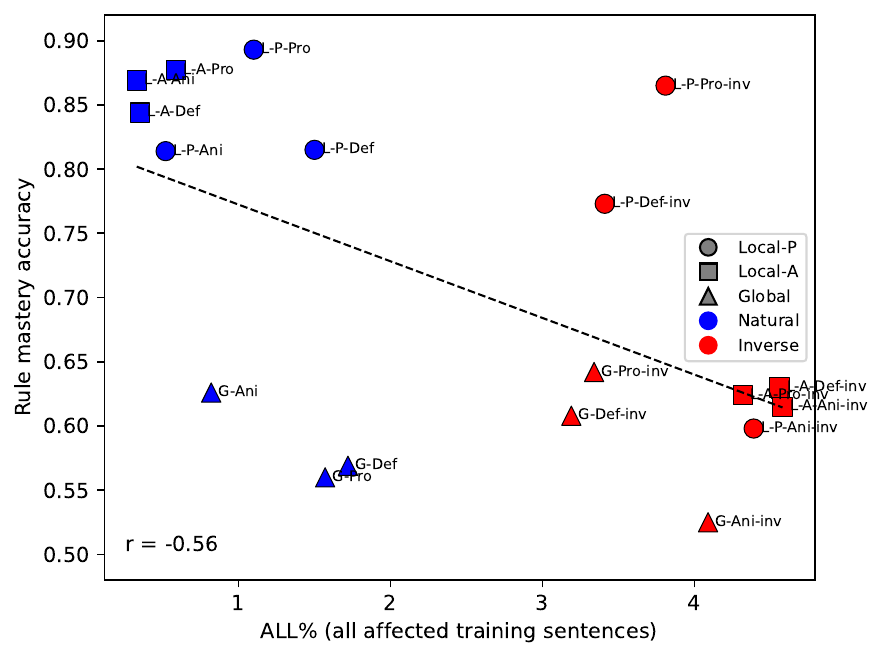}
  \caption{Scatter plot of the perturbation ratio in the training corpus (ALL\%) versus best rule mastery accuracy across DAM rules. The dashed line shows the linear regression fit.}
  \label{fig:performance-ratio-plot}
\end{figure}

\paragraph{Perturbed Ratio and Rule Mastery Accuracy.}
\label{para:freq}
To examine whether differences in DAM rule mastery can be explained primarily by input frequency, we analyze model performance as a function of perturbed ratio in the training corpus (ALL\% in Table~\ref{tab:conditions}), as shown in Figure~\ref{fig:performance-ratio-plot}. Across all rules, perturbed ratio and rule mastery exhibit a moderate negative correlation ($r=-0.56$, $p=0.0166$). 

However, frequency alone does not fully account for the observed asymmetries in model performance. While \emph{natural} rules (blue) tend to cluster around higher performance and lower perturbation ratios and \emph{inverse} rules (red) around lower performance and higher perturbation ratios, substantial variation remains within each group. Within the subset of high-performing \emph{Local--Natural} rules, the correlation between perturbed ratio and accuracy largely disappears ($r=-0.17$, $p=0.7536$), suggesting that lower perturbation ratios do not systematically lead to better rule mastery.

At the same time, markedness direction and frequency are inherently linked in naturalistic distributions: overt marking in natural languages is typically associated with less frequent or less expected argument configurations. As a result, manipulating the proportion of marked and unmarked instances also changes the typological properties of the system itself. We therefore treat frequency as a relevant factor in rule mastery, but not as a sufficient explanation for the observed pattern.

\subsection{Marker Placement Test}
\label{sec:marker-placement}

This experiment tests whether poor performance on certain DAM rules arises from a failure to learn where markers should be placed.

For each DAM condition, we construct minimal pairs from the test set. In each pair, the \emph{good} sentence contains a case marker that is licensed under the target rule given the sentence’s semantic properties, and the marker is inserted at the corresponding NP boundary. The \emph{bad} sentence is then derived by randomly shifting one required marker left or right by 1--2 tokens. For global rules that license multiple markers, we randomly select one marker to perturb. An example minimal pair is shown below:

\begin{itemize}
\item \textbf{Sentence good}: The doctor helped the boy \Ptag{}.
\item \textbf{Sentence bad}:  The doctor helped the \Ptag{} boy.
\end{itemize}

\begin{figure}[t]
    \centering
    \includegraphics[width=\linewidth]{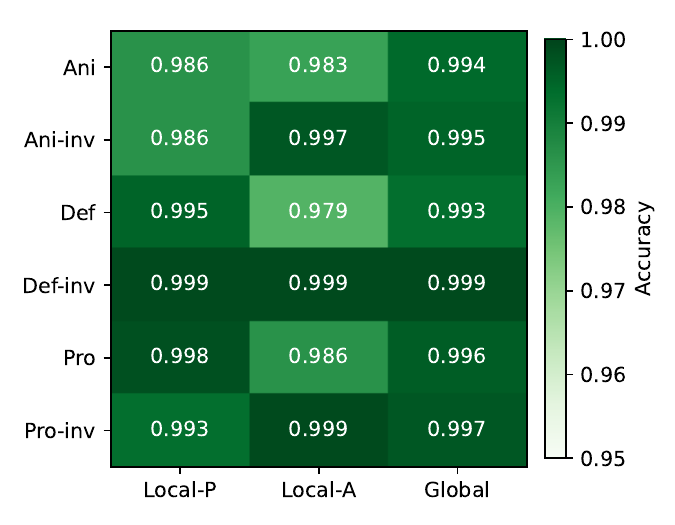}
    \caption{Marker placement accuracy of the best checkpoint across DAM conditions. We select models based on best validation perplexity.}
    \label{fig:marker-placement}
\end{figure}

We evaluate marker placement using the same minimal-pair evaluation protocol as in Section~\ref{sec:rule-mastery}. Results are summarized in Figure~\ref{fig:marker-placement}.

Across all 18 DAM rule conditions, localization accuracy is near ceiling. In contrast to the substantial variation observed in rule mastery across conditions, models overwhelmingly prefer the correctly placed markers regardless of rule type. This result indicates that models reliably learn the syntactic placement of case markers from training data, and that failures in DAM rule mastery cannot be attributed to an inability to localize markers in the surface string.

\subsection{Additional Experiments}
\label{sec:additional}

Beyond rule mastery and marker placement, we conduct two auxiliary experiments to rule out alternative explanations for differences in DAM learning (details in Appendix~\ref{app:additional}). 
First, a \emph{semantic probing} analysis tests whether rule-mastery differences reflect semantic information loss. Linear probes recover animacy, definiteness, and pronominality from subject and object representations, with no evidence of semantic degradation. 
Second, \textit{BLiMP} \citep{warstadt-etal-2020-blimp-benchmark} \textit{diagnostic tasks} show that injection of DAM does not significantly influence the learnability of other grammatical phenomena.

\section{Conclusion and Discussion}
\label{sec:conclusion-discussion}
We discuss two DAM typological tendencies in GPT-2-small models: (i) markedness direction, i.e., whether marking occurs on less usual arguments, and (ii) argument preference, i.e.\ whether marking targets the subject or object. For markedness direction, we find a clear and consistent advantage for natural systems over inverse systems, aligning with DAM typological tendencies in human language. For argument preference, object-targeting rules show only a weak overall advantage over subject-targeting rules, which does not align with the typological tendency.

One potential account of the markedness asymmetry lies in the training objective of autoregressive language models, which optimize next-token prediction. Prior work shows that such models learn more successfully when surprisal can be reduced by local contextual cues, reflecting a bias toward information locality and low local uncertainty \citep{FutrellLossy-Context2020, hahn_modeling_2021, kallini-etal-2024-mission, someya-etal-2025-information}. As less predictable meanings cause overt grammatical encoding \citep{KURUMADA2019103953}, natural markedness systems, in turn, overtly encode less usual, higher-surprisal arguments. By introducing explicit local signals that reduce prediction uncertainty, natural markedness aligns with locality-sensitive learning dynamics, offering a possible explanation for the observed advantage of natural over inverse systems.

By contrast, asymmetries in argument preference are unlikely to be driven by local advantages in next-token prediction. As discussed in Section~\ref{sec:background:argument}, the typological dominance of object-targeting DAM is commonly attributed to discourse-level and diachronic pressures. However, autoregressive models, while effective at exploiting local distributional cues, do not robustly maintain discourse-level representations \citep{kim-schuster-2023-entity, mahowald_dissociating_2024}. Consequently, they are not directly optimized for the pressures that disfavor differential subject marking, suggesting that argument-preference asymmetries depend more on discourse and historical shaping than on learnability under standard next-token objectives.

Because our corpora are constructed by injecting DAM rules into English text, distributional and morphosyntactic cues from the underlying corpus may also shape model behavior. For markedness direction, English already contains broader markedness patterns, such as unmarked singulars versus overtly marked plurals. The advantage of natural DAM rules may therefore reflect their consistency with this broader markedness direction, rather than a purely DAM-specific preference. For argument preference, English retains subject--verb agreement as a head-marking cue, which may support learning subject-targeting DAM rules. These residual cues could in principle affect model preferences if they were controlled or removed. We treat this as an important limitation, reflecting a broader constraint of the current corpus perturbation methodology.



It is important to note that while markedness and argument preference may be influenced by different mechanisms, these mechanisms are not mutually exclusive, nor are they the sole cause for an LM typological tendency. Prior work shows that markedness can emerge through communicative pressures even from initially random case-marking systems  \citep{fedzechkina2012language, fedzechkina2017balancing, smith2025communicative, lian2025simulating}. Yet our models, trained solely with a standard next-token prediction objective, still display typologically consistent markedness patterns. This suggests that markedness cannot be reduced to communicative efficiency or predictability alone, but instead reflects the joint influence of learnability and communicative pressures.

Overall, our results suggest that typological asymmetries are unlikely to be explained by a single mechanism, but instead reflect the interaction of constraints operating at different levels. LM-based experiments are valuable in this context because they provide a relatively controlled environment for isolating and testing different mechanisms from which typological tendencies may emerge. Recent work has used LMs to investigate synchronic differences in the learnability of grammatical systems, emphasizing the role of learnability in shaping typological biases \citep{kallini-etal-2024-mission, xu2025languagemodelslearntypologically}. Extending this line of research, future work could incorporate interactional or diachronic dimensions to further disentangle the relative contributions of learning mechanisms and functional pressures to typological asymmetries.

\section{Limitations}
\label{sec:limitations}

Our experiments are conducted with a single small autoregressive architecture, GPT-2-small, and on a fixed data scale, leaving open whether the observed patterns persist across model families, larger model sizes, or different training regimes. We also report results from a single random seed; evaluating across multiple seeds would provide a more reliable estimate of performance. Beyond this, we also see three methodological aspects that are worth improving.

First, our experiments are based on English, a fixed word order SVO language in which argument roles are largely recoverable from linear order and agreement alone. By contrast, typological and acquisition research has shown that rich case-marking systems are especially common in SOV and flexible word order languages, where subjects and objects often appear adjacent \citep{greenberg1963universals, vanpatten2019wordorder}. In such systems, case marking helps distinguish competing arguments more reliably. Therefore, an English-based corpus may provide less evidence for phenomena related to case marking.

Second, we perturb only clauses with a single finite verb, one subject, and one nominal direct object, leaving out ditransitives, passives, raising and control structures, and other construction types. However, in natural languages, case marking often interacts with a broader range of argument-structure alternations. Future work could extend the paradigm to a richer and more structurally diverse set of clause types.

Third, our implementation captures only a simplified version of the DAM design space. We operationalize animacy, definiteness, and pronominality as separate binary triggers, but these dimensions are not fully independent in natural languages. For example, work on Spanish DOM argues that animacy is the dominant trigger of object marking, while apparent specificity effects are closely tied to topicality and information structure \citep{Leonetti2004DOMSpanish}. On the morphological side, our implementation uses a single independent marker at the right edge of each marked NP, abstracting away from multiple coding devices, morphologically integrated case marking, and marking on nominal heads \citep{Seržant2018}. Our setup also treats triggers as sentence-internal properties, leaving aside discourse-sensitive DAM systems where marking depends on information structure, topic continuity, or context-dependent shifts in prominence or animacy.

Finally, our corpus construction strategy prioritizes preserving the overall distribution of argument structures and semantic properties present in the original data. This design choice allows us to study DAM learning under relatively naturalistic input distributions. However, it also means that the triggering frequency of different DAM rules is not experimentally controlled. As a result, while we analyze the relationship between input frequency and rule mastery, the current design cannot fully disentangle frequency effects from structural learning constraints, as discussed in Section~\ref{para:freq}. Future work could address this limitation by explicitly controlling the frequency of different SVO configurations, enabling a more precise separation of frequency-driven effects from inherent learnability differences across rules.

\bibliography{custom}
\appendix

\section{DAM Injection}
\label{app:injection}

\subsection{Corpora and Splits} \label{app:injection:corpora}
We construct the base English corpus from a subset of the OpenSubtitles dataset \citep{lison-tiedemann-2016-opensubtitles2016} (English side of the EN--FR OPUS release). 
As a light preprocessing step, we apply regex-based sentence splitting and filter sentences by length, retaining only those with 3--30 whitespace-delimited tokens. 
The resulting unperturbed corpus, prior to DAM injection, contains approximately 184M tokens and 21M sentences.

We select this dataset because its relatively well-formed conversational text supports more reliable automatic parsing and higher coverage for extracting transitive SVO clauses. 
The corpus is split into training, validation, and test sets with a 90/5/5 ratio, and these splits are defined before DAM injection and reused across all grammatical conditions to ensure strict comparability.

\subsection{Parsing and SVO Extraction} \label{app:injection:parsing}
Each sentence is parsed using spaCy \citep{honnibal-montani-2017-spacy}, augmented with Benepar constituency parsing \citep{kitaev-klein-2018-constituency}. For each sentence, we identify verbal predicates and extract clause-local predicate--argument frames by considering only dependents that are directly licensed by the predicate head. Each frame consists of one predicate together with at most one subject and one object-like argument per clause. Instances with multiple object candidates are excluded.

Argument spans are constructed by expanding noun-phrase heads into contiguous surface realizations, incorporating determiners, adjectival modifiers, compounds, and possessive elements. 

In addition to bare nominal objects, we treat certain prepositional constructions as pseudo-objects when they function as predicate-selected complements. Operationally, we identify these cases as prepositional dependents attached to the verbal predicate whose complement forms a patient/theme-like argument. Such constructions are grouped into a single argumental unit spanning the preposition and its nominal complement. This design reflects a typological and constructional property of English, in which patient-like roles are frequently encoded via prepositional marking (e.g., \emph{wait for the bus}, \emph{listen to the story}) rather than bare object positions. Cross-linguistically, such prepositional realizations often correspond to bare objects in languages with richer case morphology, motivating their inclusion in our case-marking framework \citep{blake2001case}.

\subsection{Semantic Trigger Annotation} \label{app:injection:annotation}
From the validation split, we sample subject--object pairs and collect approximately 2k NP instances (balanced across subjects and objects), covering animacy, definiteness, and pronominality. We use the GPT-4o API \citep{openai2024gpt4ocard} to generate single-token pre-labels for each NP in context, using task-specific prompts (animate/inanimate; definite/indefinite; pronoun/common). All automatically generated labels are subsequently human-verified, with independent double annotation by authors. Disagreements are resolved through discussion and adjudication, following established practices \citep{tan-etal-2024-large}.

We train three separate BERT-base classifiers \citep{devlin-etal-2019-bert}, one for each semantic trigger, on the verified seed data.
The input to each classifier consists of the full sentence, followed by a special delimiter and the target noun phrase (e.g., \texttt{The dog chased the cat [NP] the cat}), and the output is a binary label indicating the corresponding semantic property (e.g., \textsc{animate} vs.\ \textsc{inanimate}).
All classifiers are fine-tuned from \texttt{bert-base-uncased} for 10 epochs using AdamW (learning rate $2\times10^{-5}$), with held-out evaluation based on a stratified 80/20 split of the verified seed data.
On held-out portions of the seed sets, all three classifiers achieve high accuracy (approximately 97\%).
We apply the trained classifiers to the entire corpus to assign animacy, definiteness, and pronominality labels to each argument prior to applying any DAM rules.
For example, in the NP \textit{the boy}, the head \textit{boy} is labeled as [animate, definite, common].

Table~\ref{tab:pair-dists-big} and Table~\ref{tab:marginals-big} report the distributions of subject--object pairings and subject/object marginals across semantic triggers, showing the expected prominence asymmetries between subjects and objects.

\begin{table*}[t]
\centering
\small
\setlength{\tabcolsep}{6pt}
\renewcommand{\arraystretch}{1.12}
\begin{tabularx}{\textwidth}{Y r Y r Y r}
\toprule
\multicolumn{2}{c}{\textbf{Animacy pairs}} &
\multicolumn{2}{c}{\textbf{Definiteness pairs}} &
\multicolumn{2}{c}{\textbf{Pronominality pairs}} \\
\cmidrule(lr){1-2}\cmidrule(lr){3-4}\cmidrule(lr){5-6}
\textbf{Pair} & {\textbf{Count}} &
\textbf{Pair} & {\textbf{Count}} &
\textbf{Pair} & {\textbf{Count}} \\
\midrule
animate–inanimate      & 858,545 & definite–indefinite      & 669,965 & pronoun–common     & 701,852 \\
animate–animate        & 103,571 & definite–definite        & 288,003 & common–common      & 205,098 \\
inanimate–inanimate    & 62,858 & indefinite–indefinite    & 44,862 & pronoun–pronoun    & 98,329 \\
inanimate–animate      & 5,783 & indefinite–definite      & 27,927 & common–pronoun     & 25,478 \\
\bottomrule
\end{tabularx}
\caption{Distribution of subject--object pairings for three semantic triggers.}
\label{tab:pair-dists-big}
\end{table*}

\begin{table*}[t]
\centering
\small
\setlength{\tabcolsep}{6pt}
\renewcommand{\arraystretch}{1.12}
\begin{tabularx}{\textwidth}{Y r r Y r r Y r r}
\toprule
\multicolumn{3}{c}{\textbf{Animacy}} &
\multicolumn{3}{c}{\textbf{Definiteness}} &
\multicolumn{3}{c}{\textbf{Pronominality}} \\
\cmidrule(lr){1-3}\cmidrule(lr){4-6}\cmidrule(lr){7-9}
\textbf{Category} & {\textbf{Subject}} & {\textbf{Object}} &
\textbf{Category} & {\textbf{Subject}} & {\textbf{Object}} &
\textbf{Category} & {\textbf{Subject}} & {\textbf{Object}} \\
\midrule
animate      & 962,116 & 109,354 & definite      & 957,968 & 315,930 & pronoun    & 906,950 & 230,576 \\
inanimate    & 68,641 & 921,403 & indefinite    & 72,789 & 714,827 & common     & 123,807 & 800,181 \\
\bottomrule
\end{tabularx}
\caption{Marginal distributions for subjects and objects across the three semantic triggers.}
\label{tab:marginals-big}
\end{table*}

\subsection{Rule Injection and Dataset Construction}
\label{app:injection:injection}

For each sentence containing at least one predicate with a single subject and a single object, we apply the DAM rules defined in Section~\ref{sec:design} to determine whether to insert case markers. We introduce two marker symbols for this purpose: an agent marker \Atag{}, which targets the subject argument, and a patient marker \Ptag{}, which targets the object argument.

In addition to the 18 experimental conditions, we include two control settings for comparison: (i) an unperturbed baseline trained on the original corpus, and (ii) a fully perturbed condition that inserts both agent and patient markers on every eligible S--V--O frame irrespective of licensing. Each marker is inserted at the right edge of the recorded NP span, preserving all original tokens and punctuation.

Each processed sentence is assigned to exactly one of three buckets:
\begin{itemize}
  \item \textbf{Affected}: the sentence contains at least one predicate--argument frame for which the DAM rule licenses and inserts one or more markers.
  \item \textbf{Unaffected}: the sentence contains at least one valid predicate--argument frame, but no marker is licensed by the rule.
  \item \textbf{Invalid}: the sentence contains no predicate--argument frame satisfying the structural criteria (e.g., missing subject or object, clausal object, or multiple objects). All conditions share the same \emph{Invalid} set.
\end{itemize}

Sentences containing valid S--V--O frames (\emph{Affected} $+$ \emph{Unaffected}) constitute approximately 5\% of all sentences in the raw corpus. We treat \emph{Affected} and \emph{Unaffected} sentences as positive training signals for learning the DAM rule, while \emph{Invalid} sentences serve as background material that preserves the overall distribution of English outside the scope of DAM.

\subsection{Human Inspection}

To assess the reliability of the perturbation pipeline, we conduct a targeted human inspection on a small set of representative DAM rules. Specifically, we inspect \emph{L-P-Def}, \emph{L-A-Pro-inv}, and \emph{G-Ani}, which together cover different dimensions of the design space. For each rule, we manually examine 50 \emph{affected} and 50 \emph{unaffected} sentences. In addition, we inspect 100 sentences sampled from the shared \emph{Invalid} pool. The results are summarized in Table~\ref{tab:human-inspection}.

\begin{table}[t]
\centering
\small
\setlength{\tabcolsep}{6pt}
\renewcommand{\arraystretch}{1.12}
\begin{tabularx}{\columnwidth}{l l r r r}
\toprule
\textbf{Rule} & \textbf{Set} & {\textbf{$N$}} & {\textbf{Correct}} & {\textbf{Accuracy}} \\
\midrule
L-P-Def      & affected   & 50  & 47  & 94.00\% \\
L-P-Def       & unaffected & 50  & 49  & 98.00\% \\
L-A-Pro-inv   & affected   & 50  & 48  & 96.00\% \\
L-A-Pro-inv   & unaffected & 50  & 50  & 100.00\% \\
G-Ani       & affected   & 50  & 49  & 98.00\% \\
G-Ani        & unaffected & 50  & 50  & 100.00\% \\
--       & invalid    & 100 & 92  & 92.00\% \\
\midrule
\textbf{Total}       & --         & \textbf{400} & \textbf{385} & \textbf{96.25\%} \\
\bottomrule
\end{tabularx}
\caption{Results of human inspection for representative DAM rules and the \emph{Invalid} set.}
\label{tab:human-inspection}
\end{table}

Overall, the pipeline performs reliably with respect to the intended perturbation rules. For \emph{affected} and \emph{unaffected} sentences, the remaining errors are largely attributable to coordinated noun phrases. For example, in ``We do all the cooking \Ptag{} and cleaning'', the marker is not placed at the right edge of the full coordinated object. Additionally, coordinated arguments such as ``you and Emma'', in which the conjuncts have different semantic properties, are not explicitly defined in our implementation.

For the \emph{invalid} set, most errors arise from failures in extracting an underlying transitive frame. These include cases involving sentence-initial vocative constructions (e.g., ``Vincent, you protected me''), yes--no questions (e.g. ``Rygel, are you hearing this?''), and subordinate constructions (e.g. ``I know people that swear by it''), where a predicate--argument structure is present but not successfully captured by the extraction pipeline.

\section{Model Training Details}
\label{app:training}

For each DAM rule described in Section~\ref{sec:design}, we train a separate GPT-2-small language model from random initialization on the corresponding synthetic corpus. All models are trained under matched architectures and optimization settings across conditions.

Models are trained using a standard causal language modeling objective, without any explicit supervision or rule-specific signals beyond token prediction. Training data consist of the full rule-injected corpus stream, including all sentences from the \emph{Affected}, \emph{Unaffected}, and \emph{Invalid} sets for each condition.

We use the standard GPT-2 tokenizer and extend the vocabulary with two special marker symbols \Atag{} and \Ptag{}. The embedding matrix is resized accordingly. No other vocabulary items are added or modified.

All models are trained for a fixed budget of 15{,}000 optimization steps. Model checkpoints are saved every 500 steps during the first 8{,}000 training steps and every 1{,}000 steps thereafter.

Key training hyperparameters shared across all model runs are summarized in Table~\ref{tab:training-hparams}.

\begin{table}[t]
\centering
\setlength{\tabcolsep}{14pt}
\small
\begin{tabular}{lr}
\toprule
\textbf{Hyperparameter} & \textbf{Value} \\
\midrule
Model & GPT-2-small \\
Context length & 1024 \\
Training steps & 15k \\
Epochs & $\sim$8.0--8.2\footnotemark \\
Optimizer & AdamW \\
Learning rate & $3\times10^{-4}$ \\
LR schedule & Cosine (5\% warmup) \\
Batch size & 48 $\times$ 2 \\
Precision & bf16 \\
Checkpointing & 500 / 1000 steps \\
Compute & NVIDIA A100 ($\sim$6h) \\
\bottomrule
\end{tabular}
\caption{Training hyperparameters shared across all LM runs.}
\label{tab:training-hparams}
\end{table}

\section{Additional Experiments}
\label{app:additional}

\subsection{Semantic Probing}
\label{app:semantic-probing}

We conduct a semantic probing experiment to test whether argument-level semantic properties are linearly recoverable from model representations using standard probing techniques \citep{veldhoen2016diagnostic, hewitt-liang-2019-designing, belinkov-2022-probing}. We focus on three binary features underlying our DAM rules: \emph{animacy}, \emph{definiteness}, and \emph{pronominality}.

\footnotetext{Models are trained for a fixed number of optimization steps. The effective number of epochs varies slightly across DAM rules due to differences in marker density.}

For each model, probing is performed only on the best training checkpoint based on validation perplexity. Probing sentences are drawn from the annotated test data. For each sentence, we extract the representation of the argument head (subject or object) by selecting the final-layer hidden state of the rightmost token whose character span overlaps the annotated head span.

\begin{figure*}[t]
    \centering
    \includegraphics[width=\linewidth]{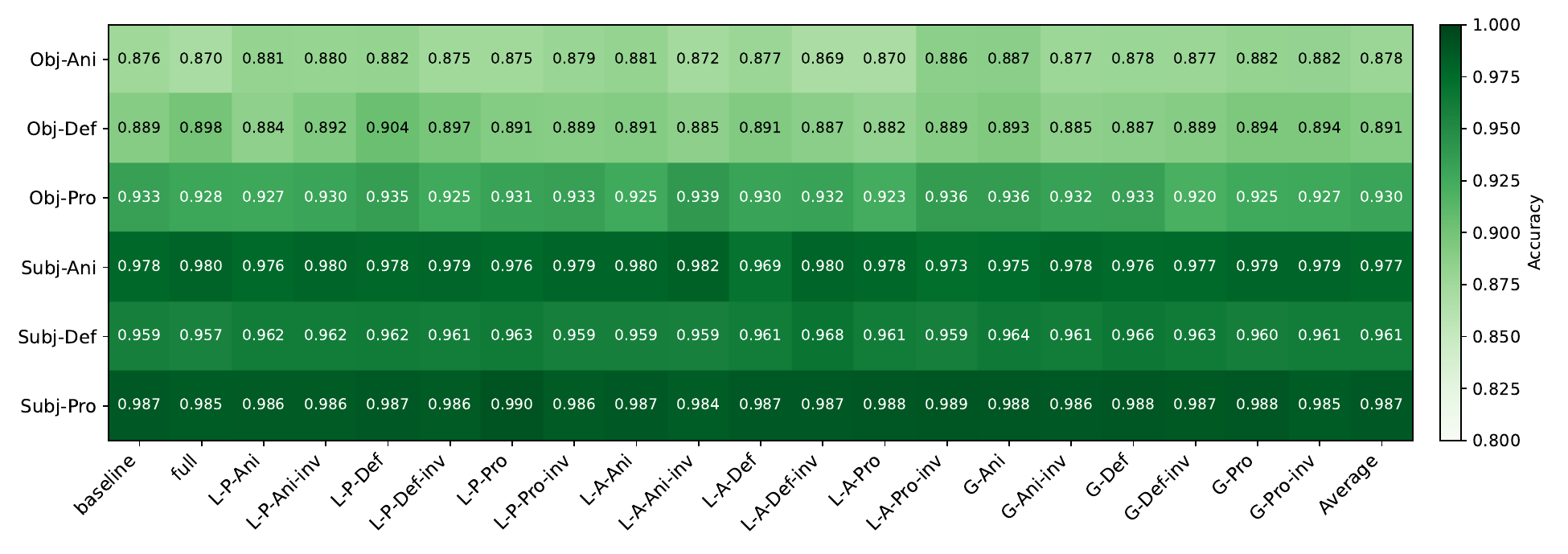}
    \caption{Semantic probing accuracy at the best training checkpoint across DAM conditions, evaluated separately for subject and object representations.}
    \label{fig:semantic-probing}
\end{figure*}

For each feature, we train a separate binary linear classifier on top of the extracted representations, with probes trained independently for subjects and objects. Balanced datasets are constructed by sampling equal numbers of positive and negative instances, with 2{,}000 examples per class for training and 1{,}000 per class for testing. We report classification accuracy on the balanced test sets in Figure~\ref{fig:semantic-probing}.

Across all rules, models show high linear separability for all three semantic features, with no systematic degradation across rule conditions. Subject representations consistently yield higher probing accuracy than object representations. This asymmetry aligns with the rule mastery results, where the performance gap between \emph{natural} and \emph{inverse} variants is smaller for \emph{Local-P} rules than for \emph{Local-A} rules.

Overall, these results indicate that differences in DAM rule mastery across conditions cannot be attributed to a failure to learn or encode semantic information.

\subsection{BLiMP Evaluation}
\label{app:blimp}

To assess whether DAM perturbations interfere with broader grammatical learning, we evaluate trained models on a subset of BLiMP benchmarks \citep{warstadt-etal-2020-blimp-benchmark}. We select eight BLiMP sub-tasks spanning core syntactic and syntax--semantics phenomena: \emph{Determiner--Noun Agreement}, \emph{Subject--Verb Agreement}, \emph{NPI Licensing}, \emph{Existential There Quantifiers}, \emph{Animate Subject Passive}, \emph{Animate Subject Transitive}, \emph{Transitive}, and \emph{Intransitive}.

To ensure distributional consistency between training and evaluation, we additionally construct DAM-perturbed versions of each selected BLiMP sub-task. In these variants, argument markers are inserted or retained according to the corresponding DAM rule, without altering the grammatical labels of the original BLiMP items.

\begin{figure*}[t]
    \centering
    \includegraphics[width=\linewidth]{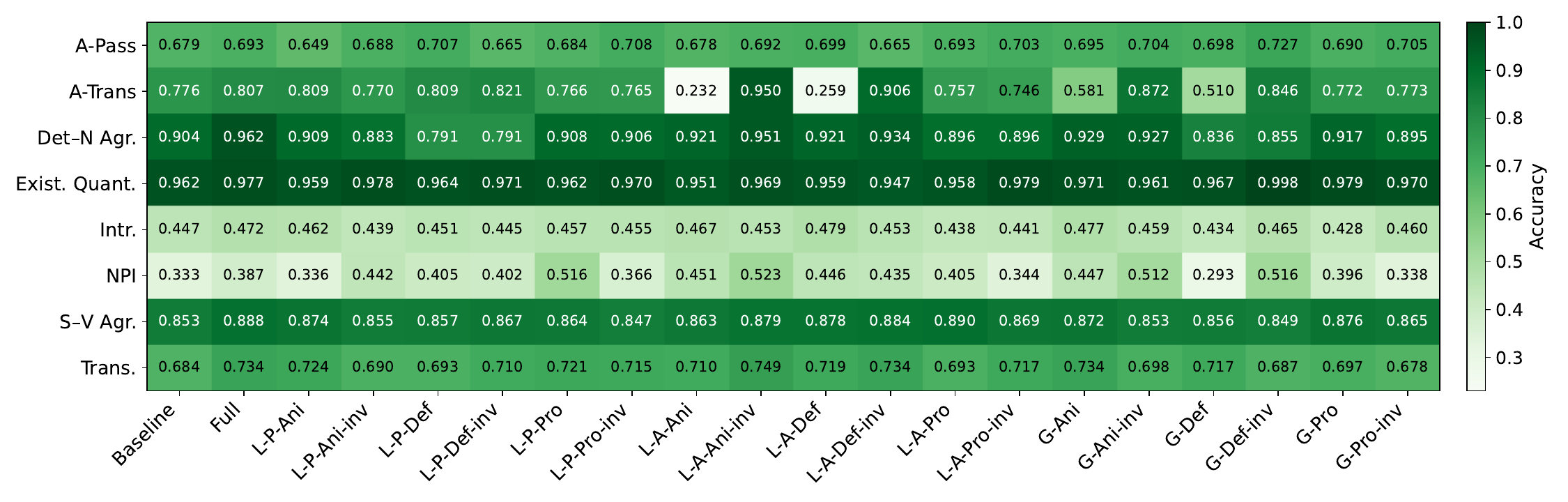}
    \caption{Full BLiMP evaluation results across all DAM rules and the baseline. 
Column abbreviations: A-Pass = Animate Subject Passive; A-Trans = Animate Subject Transitive; 
Det-N Agr. = Determiner–Noun Agreement; Exist. Quant. = Existential Quantifiers; 
Intr. = Intransitive; NPI = Negative Polarity Item Licensing; 
S–V Agr. = Subject–Verb Agreement; Trans. = Transitive.}
    \label{fig:blimp}
\end{figure*}

\begin{figure}[t]
    \centering
    \includegraphics[width=\linewidth]{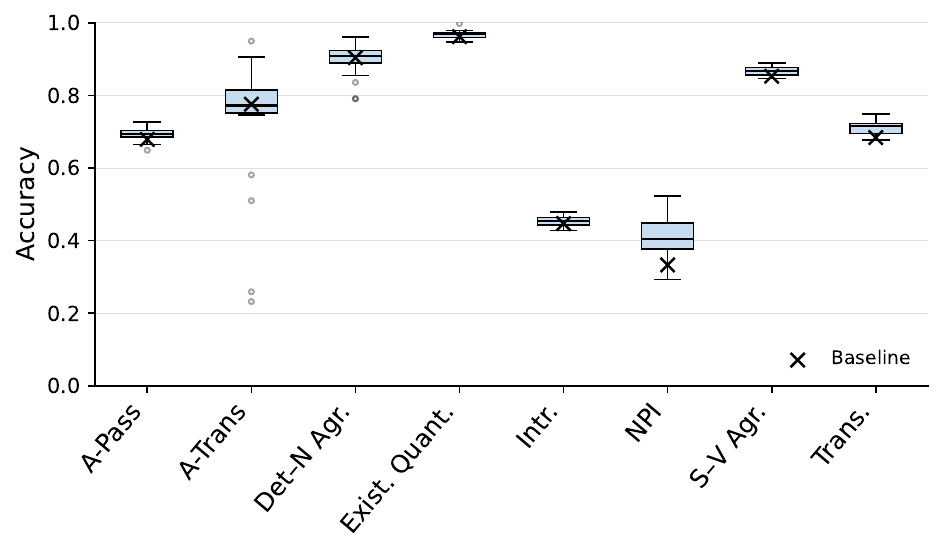}
    \caption{BLiMP accuracies across eight sub-tasks. 
  Boxplots show the distribution across DAM rules; black "×" markers denote the baseline model.}
    \label{fig:blimp_boxplot}
\end{figure}

We evaluate each model (including the \emph{baseline} and \emph{full-perturbation} controls) on DAM-perturbed BLiMP sets using a likelihood-based minimal-pair evaluation, as in Section~\ref{sec:rule-mastery}. We report sub-task accuracy at the best training checkpoint in Figure~\ref{fig:blimp} and summarize accuracy distributions across rules in Figure~\ref{fig:blimp_boxplot}.

Across the eight BLiMP sub-tasks, DAM-perturbed models show accuracy distributions tightly centered around the unperturbed baseline, with no consistent downward shift \footnote{An exception arises in the \textit{Animate Subject Transitive} subtask, which rewards analyses that prefer animate transitive subjects. Rules introducing markers on animate agents, such as \emph{L–A–Ani-inv}, may therefore align more closely with BLiMP’s grammatical variants (e.g., \emph{Beth \Atag{} scares Roger}). As a result, observed accuracy differences on this sub-task reflect task--rule alignment effects rather than improvements in broader grammatical competence.}. Overall, these results indicate that DAM perturbations do not interfere with broader grammatical learning beyond the targeted argument-marking behavior.

\end{document}